\documentclass[10pt,twocolumn,letterpaper]{article}

\usepackage{iccv}
\usepackage{times}
\usepackage{epsfig}
\usepackage{graphicx}
\usepackage{amsmath}
\usepackage{amssymb}

\usepackage[breaklinks=true,bookmarks=false]{hyperref}

\iccvfinalcopy 

\def\iccvPaperID{****} 

\ificcvfinal\fi

\begin{document}

\title{An End-to-End Trainable Video Panoptic Segmentation Method using Transformers}

\author{Jeongwon Ryu~~~and~~~Kwangjin Yoon\thanks{Kwangjin Yoon is a corresponding author.} \\
SI Analytics\\
Yuseong-daero 1689, Yuseong-gu, Daejeon, 34047, Republic of Korea\\
{\tt\small \{rjw0926;yoon28\}@si-analytics.ai}
}

\maketitle
\ificcvfinal\thispagestyle{empty}\fi

\begin{abstract}
In this paper, we present an algorithm to tackle a video panoptic segmentation problem, a newly emerging area of research. The video panoptic segmentation is a task that unifies the typical task of panoptic segmentation and multi-object tracking. In other words, it requires generating the instance tracking IDs along with panoptic segmentation results across video sequences. Our proposed video panoptic segmentation algorithm uses the transformer and it can be trained in end-to-end with an input of multiple video frames. We test our method on the STEP dataset and report its performance with recently proposed STQ metric. The method archived 57.81\% on the KITTI-STEP dataset and 31.8\% on the MOTChallenge-STEP dataset.
\end{abstract}

\section{Introduction}

Comprehensive video understanding can be efficiently accomplished with the video panoptic segmentation, since it simultaneously tackles multiple tasks, namely segmentation of the scene elements and identification of the instances. 

In this paper, we present a transformer based video panoptic segmentation method, named SIAin. The model is built from MaskFomer \cite{cheng2021maskformer}, a recently proposed panoptic segmentation algorithm. MaskFomer formulates segmentation task as a mask classification instead of a per-pixel classification \cite{cheng2021maskformer}. However, since MaskFormer merely processes each image independently in a video sequence, we modified it to have the object tracking functionality. The proposed method is trained with a new loss function which consists of detection-loss and tracking-loss. The detection-loss is minimized if newly appeared objects and semantic objects are correctly segmented while the tracking-loss minimizes object tracking errors. Both detection and tracking losses are formulated similarly with \cite{cheng2021maskformer}.

We take SIAin to participate in the video track competition of Benchmarking Multi-Target Tracking (BMTT) workshop 2021. The competition requires to assign semantic classes and track identities to all pixels in a video. The competition provides a new dataset, STEP \cite{weber2021step}. The performance of each entry is measured with the STQ \cite{weber2021step} metric. Our method archived 57.81\% on the KITTI-STEP dataset and 31.8\% on the MOTChallenge-STEP dataset.

\section{Method}

\begin{figure*}[t]
    \centering
    \includegraphics[width=0.95\textwidth]{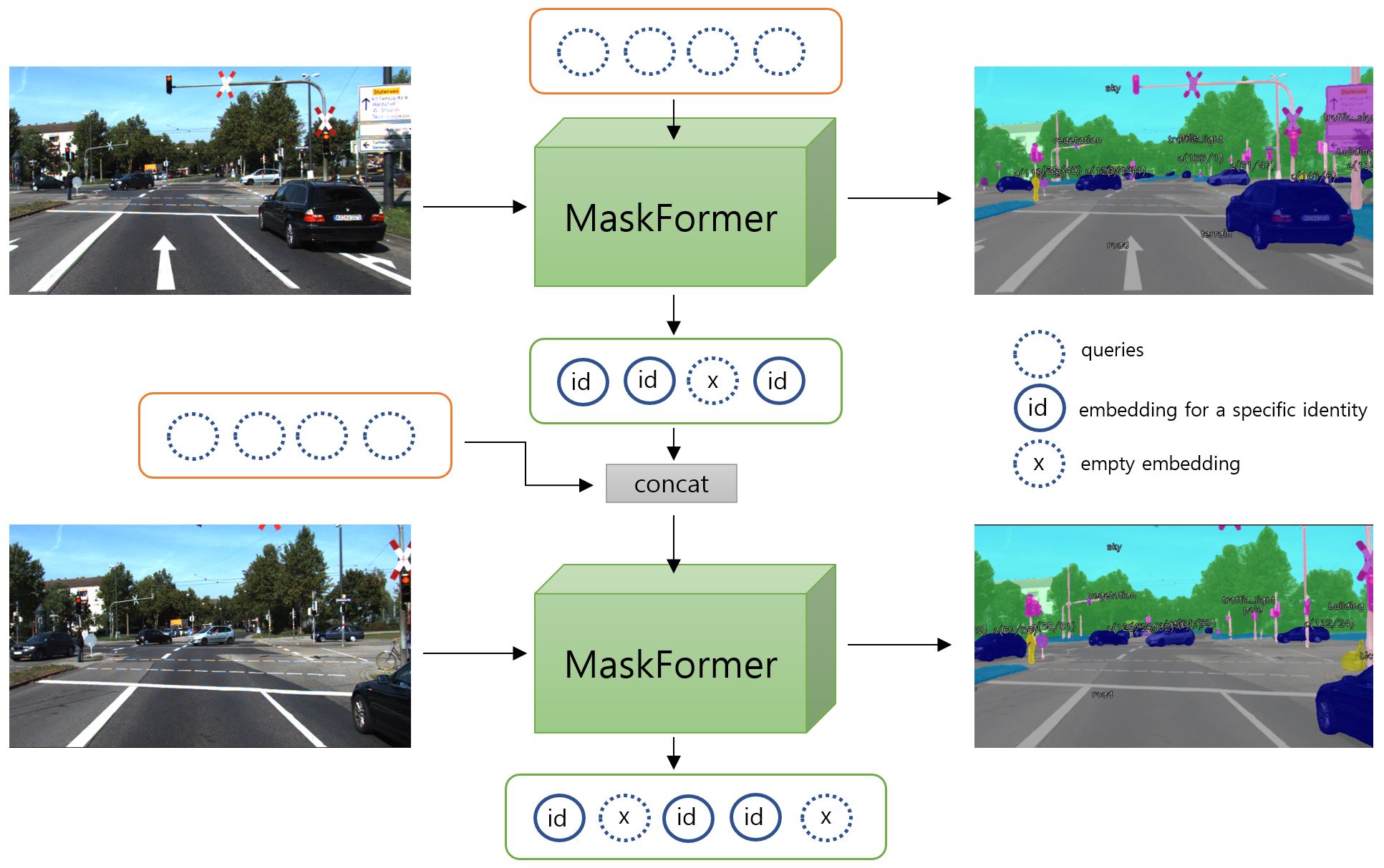}
    \caption{Brief description of the proposed architecture}
    \label{fig:arch}
\end{figure*}

We adopt MaskFomer \cite{cheng2021maskformer} to build a video panoptic segmentation network. MaskFomer can generate the panoptic segmentation result for an image by formulating segmentation task as a mask classification task instead of a per-pixel classification \cite{cheng2021maskformer}. In order to make MaskFormer have the tracking functionality, we mainly modified its loss function. At each iteration of training phase, ours network accepts $K$ frames instead of a single image. We call this $K$ training sequences as \textit{episode}. In an episode, there are a set of semantic objects ($S$), a set of detected objects ($D$), and a set of tracked objects ($T$). A semantic object is an object that belongs to semantic categories, such as sky, building, sidewalk, vegetation and so on. Both detected objects and tracked objects have a class required to be tracked by the video panoptic segmentation algorithm, such as car and person in the STEP dataset. The difference between the detected and the tracked objects is the newness of an object given a certain identity. For example, let us assume that there is an object moving (and visibly) through the episode. Then, the initial instance of the object is classified as the detected object and the remaining instances of the object are classified into the tracked object set. Using the sets $S$, $D$ and $T$, we can compute the detection loss $L_{S,D}$ and the tracking loss $L_T$. Specifically, the loss $L_{S,D}$ is computed with set $S$ and $D$, while the loss $L_T$ is computed with set $T$. Both detection and tracking losses are similarly formulated with \cite{cheng2021maskformer} except that the ground-truth masks are clustered into set $S$, $D$, and $T$. The final loss is computed as follow:
\begin{equation}
    L = \lambda_{S,D}L_{S,D} + \lambda_{T}L_{T}
\end{equation}
where $\lambda_{S,D}$ and $\lambda_T$ are weight constants for the detection loss and the tracking loss, respectively.
Furthermore, it is worth note that we insert embeddings of detected/tracked objects into the next frames' transformer decoder as queries in order to make our method efficiently track objects.

In the test phase, we use our end-to-end trained SIAin to solve video panoptic segmentation problem. At the very first frame, the initial frame of the sequence is panoptic-segmented by SIAin, and it saves the embeddings (outputs of the transformer decoder) of detected objects. Then, the embeddings are inserted to the transformer decoder in the next frame. If an object is tracked, its embedding is replaced with new one. Finally, a tracked object is terminated if the object missed in $M$ consecutive frames. 

A brief description of our proposed archtecture is depicted in Figure \ref{fig:arch}.

\subsection{Implementation Details}

The episode length $K$ is randomly sampled from a discrete uniform distribution with an interval $[2, 5]$. Then, we compose an episode with $K$ consecutive frames after choosing the initial frame and interval. The initial frame is also randomly chosen between the start and end frame with sufficient margin. The interval between consecutive frames are also randomly sampled from a discrete uniform distribution with an interval $[1, 4]$.


In the training phase, SIAin is trained on an Nvidia A100 GPU following little changes to the default settings of \cite{cheng2021maskformer}. We use Swin-L \cite{liu2021swin} for the backbone and set $\lambda_{S,D}$ and $\lambda_T$ to 0.3 and 0.7, respectively. Here, we omit the other details described in \cite{cheng2021maskformer}.
In the test phase, $M$ is set to $5$ for both KITTI-STEP and MOTChallenge-STEP datasets.

\section{Experiments}

\begin{figure*}[t]
\begin{center}
\includegraphics[width = \textwidth]{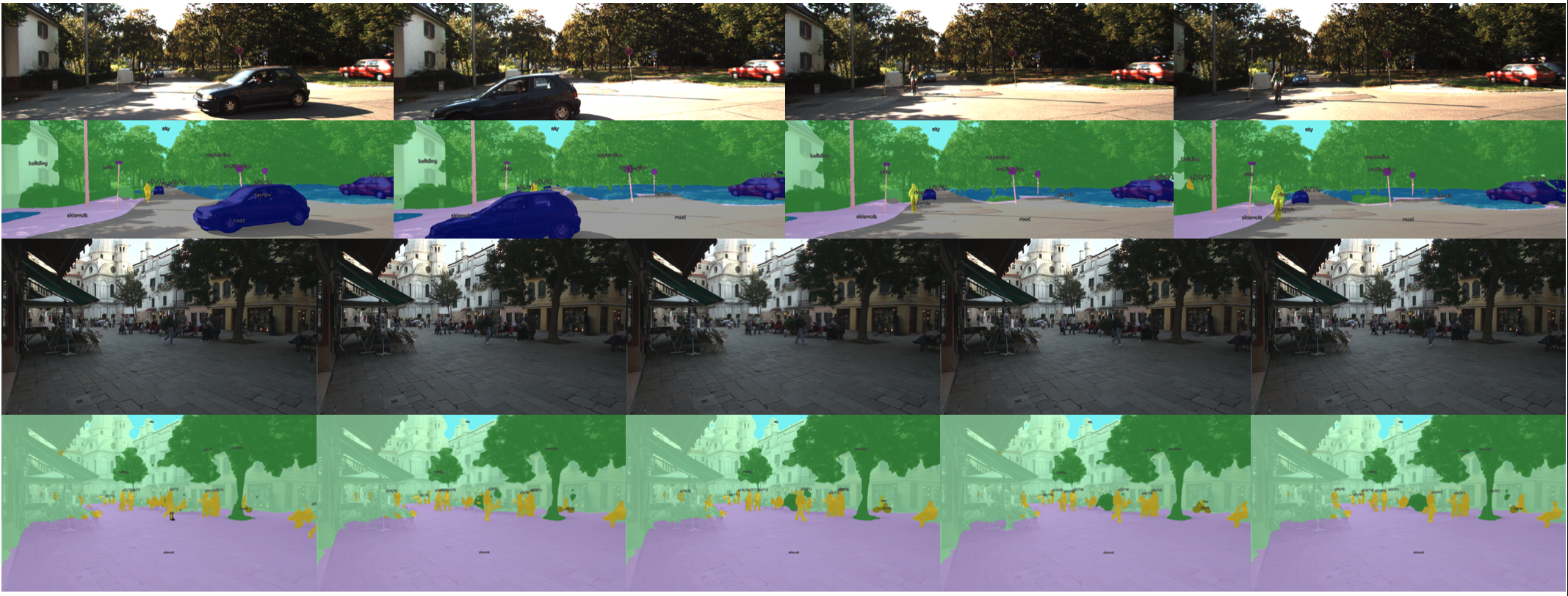}
\end{center}
   \caption{ Qualitative results of SIAin. 1st, 2nd row: KITTI-STEP; 3rd, 4th row: MOTChallenge-STEP.}
\label{fig:results}
\end{figure*}

\subsection{Datasets}

The video track of the BMTT Challenge is evaluated by Segmenting and Tracking Every Pixel (STEP) \cite{weber2021step}, a new benchmark that includes two datasets: KITTI-STEP and MOTChallenge-STEP. The MOTChallenge-STEP dataset has four sequences, and equally divided into two training and two testing sequences. It provides instance-level annotations and identities for persons class. The KITTI-STEP dataset consists of 21 training sequences and 29 test sequences. It annotates identities and instances for cars and persons classes. The semantic classes are slightly different from each other. Please, refer to \cite{weber2021step}.

\subsection{Challenge Results}

We report the performance of the SIAin. The performance of each entry at the video track of BMTT is measured by STQ metrics \cite{weber2021step}. Our method archived 57.81\% on the KITTI-STEP dataset and 31.8\% on the MOTChallenge-STEP dataset (Table \ref{table:kitti} and Table \ref{table:mots}). Some qualitative examples are shown in  Figure~\ref{fig:results}. Moreover, in table \ref{table:kitti}, our method outperform the baseline method, Motion-DeepLab \cite{weber2021step}, on the KITTI-STEP by a large margin (5.68). 


%
\setlength{\tabcolsep}{4pt}
\begin{table}[ht]
\begin{center}
\begin{tabular}{cccc}
\hline
{ {\textbf{Method}}} &   { {\textbf{STQ}}} &  {\textbf{AQ}} &  {\textbf{SQ(IoU)}} \\
\hline \hline
 Motion-DeepLab   & 52.19\% &	45.55\%	& 59.81\% \\
SIAin (ours)             & \textbf{57.87\%} & \textbf{55.16\%} & \textbf{60.71\%} \\ 
\hline
\end{tabular}
\end{center}
\caption{
 KITTI-STEP leaderboard results.
}
\label{table:kitti}
\end{table}

\setlength{\tabcolsep}{4pt}
\begin{table}[ht]
\begin{center}
\begin{tabular}{cccc}
\hline
{ {\textbf{Method}}} &   { {\textbf{STQ}}} &  {\textbf{AQ}} &  {\textbf{SQ(IoU)}} \\
\hline \hline
 SIAin (ours)             & \textbf{31.8\%} & \textbf{18.4\%} & \textbf{65.7\%} \\ 
\hline
\end{tabular}
\end{center}
\caption{
 MOTChallenge-STEP leaderboard results.
}
\label{table:mots}
\end{table}

\section{Conclusion}

We present an end-to-end framework for video panoptic segmentation using transformers. We adopted MaskFormer \cite{cheng2021maskformer} to make video panoptic segmentation algorithm, SIAin. SIAin uses query embeddings that follow objects over a sequence as an autoregressive manner. The proposed framework can effectively assign a semantic class and track ID to every pixel and can also be used in online tracking scenarios. Our method archived 57.81\% STQ on the KITTI-STEP dataset and 31.8\% STQ on the MOTChallenge-STEP dataset. 

{\small
\bibliographystyle{ieee_fullname}
\bibliography{siain}
}

\end{document}